\documentclass[twoside,11pt]{article}

\usepackage{blindtext}
\usepackage[abbrvbib, preprint]{ArXiv}
\usepackage[svgnames]{xcolor}
\hypersetup{
hidelinks,
colorlinks=true,
linkcolor=Indigo,
urlcolor=DeepSkyBlue4,
citecolor=Indigo
}
\usepackage{booktabs}
\usepackage{dsfont}
\usepackage[inline]{enumitem}
\usepackage{amsmath}
\usepackage{algorithm}
\usepackage{algorithmic}
\usepackage{subcaption}
\usepackage[english]{babel}
\usepackage{hyphenat}
\usepackage{natbib}

\newcommand{\PARAMETER}{\item[\textbf{Parameter:}]}

\usepackage{marvosym}
\makeatletter
\def\@fnsymbol#1{\ifcase#1\or \text{\Letter}\or *\or \dagger\or \ddagger\else\@arabic{#1}\fi}
\makeatother

\usepackage{lastpage}
\jmlrheading{23}{\ Preprint 2025}{1-\pageref{LastPage}}{1/21; Revised 5/22}{9/22}{21-0000}{}

\firstpageno{1}

\begin{document}

\title{Improving Multi-Label Contrastive Learning by Leveraging Label Distribution}

\author{\name Ning Chen \email che-n-ing@hhu.edu.cn\\
\addr Key Laboratory of Water Big Data Technology of Ministry of Water Resources,\\ College of Computer Science and Software Engineering, Hohai University, Nanjing, China\\
\AND
\name Shen-Huan Lyu~\thanks{Corresponding author} \email lvsh@hhu.edu.cn \\
\addr Key Laboratory of Water Big Data Technology of Ministry of Water Resources,\\ College of Computer Science and Software Engineering, Hohai University, Nanjing, China\\
\addr State Key Laboratory for Novel Software Technology, Nanjing University, Nanjing, China\\
\AND
\name Tian-Shuang Wu \email tianshuangwu@hhu.edu.cn \\
\addr Key Laboratory of Water Big Data Technology of Ministry of Water Resources,\\ College of Computer Science and Software Engineering, Hohai University, Nanjing, China\\
\AND
\name Yanyan Wang \email yanyan.wang@hhu.edu.cn \\
\addr Key Laboratory of Water Big Data Technology of Ministry of Water Resources,\\ College of Computer Science and Software Engineering, Hohai University, Nanjing, China\\
\AND
\name Bin Tang \email cstb@hhu.edu.cn \\
\addr Key Laboratory of Water Big Data Technology of Ministry of Water Resources,\\ College of Computer Science and Software Engineering, Hohai University, Nanjing, China\\
}

\editor{My editor}

\maketitle

\begin{abstract}
In multi-label learning, leveraging contrastive learning to learn better representations faces a key challenge: selecting positive and negative samples and effectively utilizing label information. Previous studies selected positive and negative samples based on the overlap between labels and used them for label-wise loss balancing. However, these methods suffer from a complex selection process and fail to account for the varying importance of different labels. To address these problems, we propose a novel method that improves multi-label contrastive learning through label distribution. Specifically, when selecting positive and negative samples, we only need to consider whether there is an intersection between labels. To model the relationships between labels, we introduce two methods to recover label distributions from logical labels, based on Radial Basis Function (RBF) and contrastive loss, respectively. We evaluate our method on nine widely used multi-label datasets, including image and vector datasets. The results demonstrate that our method outperforms state-of-the-art methods in six evaluation metrics.

\end{abstract}


\section{Introduction}
Multi-label learning aims to construct a model that can effectively assign a set (multiple) of labels associated with an instance to that instance~\citep{Zhang2014AReview}. For example, in the case of a scenic image composed of labels such as ``person'', ``waves'', ``beach'', and ``sky'', the model needs to learn these labels associated with the scenic image. Currently, multi-label learning has a wide range of applications, such as text classification~\citep{Zeng2024Multi-Label}, image annotation
~\citep{Yuan2023Graph}, gene function prediction~\citep{MinLing2006MNN}, musical instrument classification in music~\citep{Zhi2023An}, and video annotation~\citep{Markatopoulou2019Implicit}, among others.

The remarkable research in contrastive learning~\citep{He2020MoCo,Chen2020SimCLR} has injected a strong impetus into the study of multi-label learning. Indeed, multi-label learning and contrastive learning share a common goal: to extract more meaningful feature representations from data. Based on this, researchers in multi-label learning have turned their attention to contrastive learning. However, these studies~\citep{He2020MoCo,Chen2020SimCLR} focus on contrastive learning in self-supervised scenarios, while multi-label learning is typically conducted in supervised settings. Therefore, exploring supervised contrastive learning for multi-label learning has become an urgent and crucial research direction. A promising development is that the research by~\citet{Khosla2020Supervised} has extended unsupervised contrastive learning to supervised contrastive learning, achieving impressive results on the ImageNet dataset. Additionally, they proposed two supervised contrastive loss functions, which provide significant guidance for subsequent research in multi-label learning.

The core idea of contrastive learning is to learn feature representations by comparing positive and negative samples, with a key aspect being the definition and selection of these positive and negative samples~\citep{He2020MoCo,Chen2020SimCLR,Khosla2020Supervised}. Building on the work of~\citet{Khosla2020Supervised}, ~\citet{Zhang2024MulSupCon} further addressed the challenge of defining positive and negative samples in the context of multi-label learning. They proposed three solutions based on the degree of label overlap: \begin{enumerate*}[label=\arabic*)] 
    \item \textbf{ALL}: A sample qualifies as a positive sample if its labels completely overlap with those of the query sample; otherwise, it qualifies as a negative sample.
    \item \textbf{ANY}: A sample is a positive sample if its labels include any of the labels of the query sample.
    \item MulSupCon~\citep{Zeng2024Multi-Label}: Building on the \textbf{ANY} method, this method separately considers the labels of samples to balance the loss weights among different labels.
\end{enumerate*} 
Building on the work of~\citet{Zhang2014AReview}, ~\citet{Guangming2024Similarity-Dissimilarity} further proposes five types of relationships between sample labels and anchor labels, dynamically reweighting the loss based on these relationships.

\begin{figure}[t]
    \centering
    \begin{minipage}[t]{0.30\columnwidth} 
        \includegraphics[width=\textwidth]{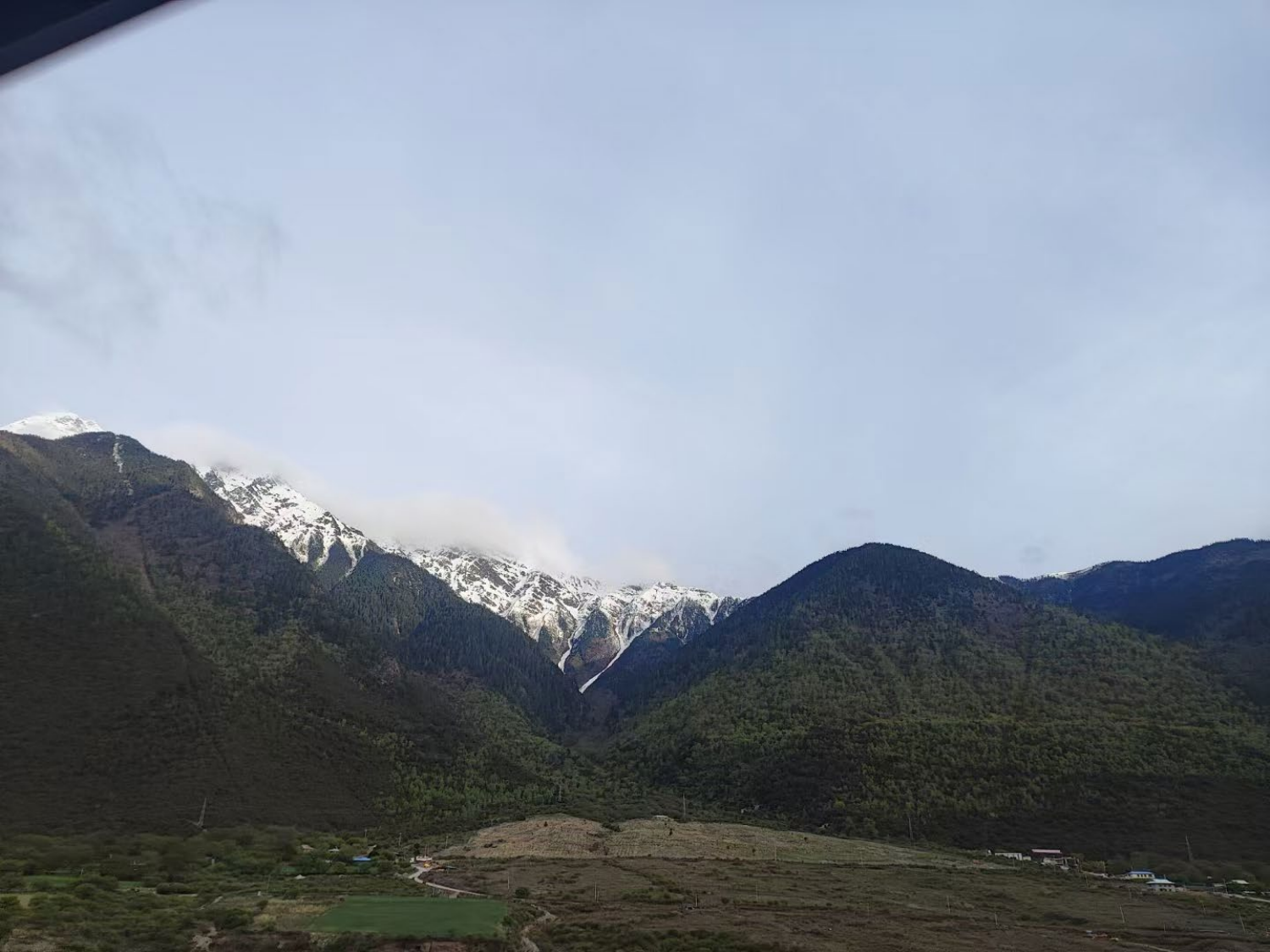}
    \end{minipage}
    \hfill
    \begin{minipage}[t]{0.315\columnwidth} 
        \includegraphics[width=\textwidth]{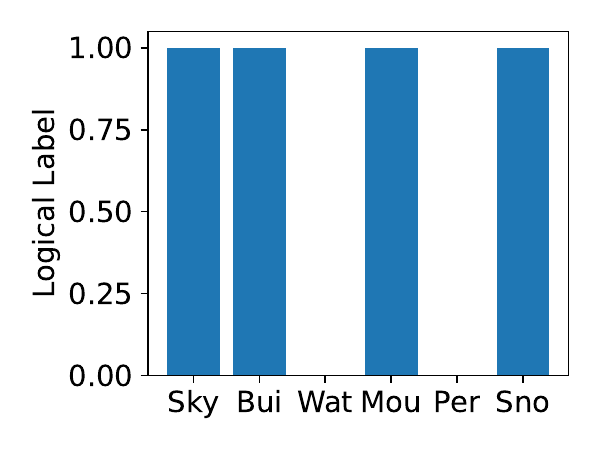}
    \end{minipage}
    \hfill
    \begin{minipage}[t]{0.315\columnwidth} 
        \includegraphics[width=\textwidth]{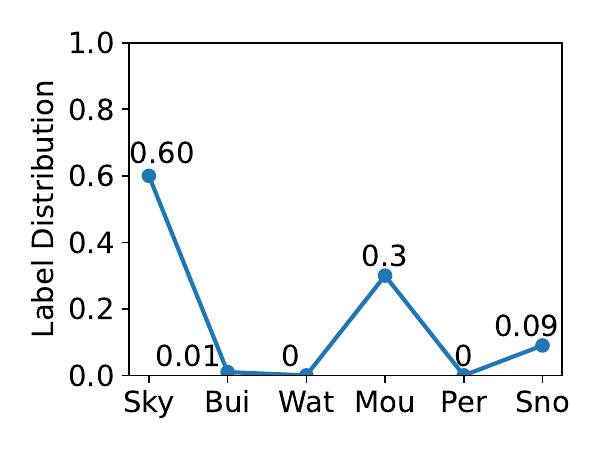}
    \end{minipage}
    \caption{Example of label distribution. Logical labels and label distributions provide different descriptions of an image. The label ``Bui'', ``Wat'', ``Mou'', ``Per'' and ``Sno'' represents ``Building'', ``Water'', ``Mountain'', ``Person'' and ``Snow'', respectively.}
    \label{fig:explanation-of-label-distribution}
\end{figure}

However, these multi-label contrastive learning assume that all labels are of equal importance, which clearly contradicts the realities of the real world. For example, the Figure~\ref{fig:explanation-of-label-distribution} shows that although some buildings exist, their importance clearly differs compared to other labels. To account for the varying importance of labels in contrastive learning, we have introduce label distribution. Unlike logical labels, which can only indicate whether a label belongs to a sample, label distribution further represents the degree of relevance or importance of a label to an instance and we can find this in Figure~\ref{fig:explanation-of-label-distribution}. This distribution can more finely characterize the ambiguity and uncertainty between instances and labels. Moreover, research by~\citet{Geng2016Label} demonstrates that label distribution contributes to better generalization performance of models. However, it is a fact that the multi-label datasets we use do not provide label distributions, making it a challenge to obtain them. Fortunately, some progress already exists in this area~\citep{Shao2018Multilabel,Ning2021LabelEnhancement,Zhang2021Leveraging,Zheng2023Generalized}. However,~\citet{Jia2023LabelDL} points out that these methods, when obtaining label distributions, may assign some degree of description to irrelevant labels. Therefore, to reduce the occurrence of such situations in our work, we enhance the method based on~\citet{Jia2023LabelDL} to make it suitable for contrastive learning.

Finally, we propose a novel method \textbf{MulSupCon$_{\text{LD}}$} that improves multi-label contrastive learning through label distribution. Firstly, to reduce the difficulty of selecting positive samples, we adopt the \textbf{ANY} strategy, meaning that if labels share any intersection, we consider the label a positive sample. Secondly, to model the relationships between labels, we recover the label distribution from the logical labels. Overall, our key contributions include the following:
\begin{enumerate}[label=\arabic*)] 
    \item To the best of our knowledge, we are the first to introduce label distribution into contrastive learning and apply it to multi-label classification tasks. We reduce the difficulty of selecting positive and negative samples and employ two methods to recover label distributions within contrastive learning.
    \item Our method not only learns more discriminative feature representations but also more effectively captures the dependencies between labels.
    \item We evaluate our method on nine commonly used multi-label datasets, including image datasets and vector datasets. The results demonstrate that our approach performs well across six metrics.
\end{enumerate}

\section{Related Work}
The core idea of contrastive learning is to learn low-dimensional representations of data by contrasting positive and negative samples, so that similar samples are close to each other in the feature space, while dissimilar samples are far apart.~\citet{Chen2020SimCLR}, in their work SimCLR, emphasize that data augmentation is a key aspect of contrastive learning, a suggestion that we adopt in our work.~\citet{He2020MoCo} propose MoCo, which tackles the problem of the number of negative samples being tied to and insufficient for the batch size and maintains the consistency of negative samples through momentum encoding, a solution we use in our work.~\citet{Khosla2020Supervised} extend the research of contrastive learning from self-supervised to supervised scenarios and use label information to leverage multiple positive samples, thereby enhancing model performance. Besides, the CCL-SC~\citep{Wu2024confidence} method optimizes the feature layer through contrastive learning, offering a new perspective for selective classification, reducing selective risk and achieving significant performance improvements.

However, these studies focus on single-label scenarios. Given the notable achievements of contrastive learning in both self-supervised and supervised scenarios, researchers in multi-label learning explore its application to multi-label learning.~\citet{Shu2022Use} propose a hierarchical multi-label representation learning framework that can fully utilize all available labels and preserve the hierarchical relationships between categories.~\citet{Zhang2024MulSupCon} introduce MulSupCon, which balances the loss based on the degree of overlap between labels.~\citet{Guangming2024Similarity-Dissimilarity}, building on the work of~\citet{Zhang2024MulSupCon}, propose five types of relationships between sample labels and anchor labels and dynamiclly balance the contrastive loss based on these five degrees of similarity and dissimilarity.

However, these studies in multi-label learning directly use logical labels, which overlooks the varying degrees of importance among labels and fails to fully exploit the relationships between labels, thereby limiting the potential to enhance model performance. Therefore, we consider incorporating label distribution into contrastive learning. However, since the datasets do not provide label distributions, to address this problem, we turn our attention to label distribution learning.
Some excellent works exist in the field of label distribution learning~\citep{Geng2016Label,Jia2021LabelDL,Jia2023LabelDistribution,Ren2019LabelDistribution,Shao2018Multilabel,Wang2023LabelDistri,Xu2017IncompleteLD}. For instance,~\citet{Geng2016Label} employes the maximum entropy model and Kullback-Leibler divergence to learn label distributions,~\citet{Wang2023LabelDistri} explores label relationships through manifolds to enhance the model's capability, and~\citet{Ren2019LabelDistribution} adopts low-rank matrix approximation to learn label distributions. 

However, their efforts largely focus on training with datasets that already possess label distributions, whereas the dataset we use lacks such distributions, and our goal is to recover the label distributions from logical labels. Furthermore, when restoring label distributions, they often assign a certain degree of description to irrelevant labels to a significant extent. 
Therefore, building on DLDL~\citep{Jia2023LabelDL}, we propose a novel method that integrates contrastive learning with label distribution learning to recover label distributions from logical labels.

\section{The Proposed Method}
To fully utilize label information in contrastive learning, we propose a novel model called \textbf{MulSupCon$_\text{LD}$}. Unlike previous multi-label contrastive learning, such as MulSupCon ~\citep{Zhang2024MulSupCon}, which directly use logical labels to balance the loss between labels, thereby overlooking the crucial point that the degree of importance among labels related to instances varies, we incorporate label distribution into contrastive learning. Label distribution~\citep{Geng2016Label} inherently reflects the varying degrees of importance among labels, rather than simply categorizing them as relevant or irrelevant, which implies that the model can learn richer information. Given that label distribution is often difficult to obtain and is absent in the datasets we use as well as in the data of comparative methods, we draw inspiration from DLDL~\citep{Jia2023LabelDL} to recover label distribution from logical labels. Subsequently, we apply the label distribution to the training process of contrastive learning. Our model will simultaneously optimize both aspects: enhancing the feature representation of samples through contrastive learning and refining label relationships through label distribution.
\paragraph{Notations}
Let $\mathbf{\mathcal{B}}=[(\mathbf{x}_1, \mathbf{y}_1);(\mathbf{x}_2, \mathbf{y}_2);\dots;(\mathbf{x}_n, \mathbf{y}_n)]$ denotes a batch of data, where $n$ signifies the size of the batch, $x_i$ represents the $i$-th sample, and $y_i$ denotes the logical label of the $i$-th sample. $\mathbf{y}_i=
[y_i^1,y_i^2,\dots, y_i^c]$, where $c$ and $y_i^j$ denotes the size of labels of the sample $i$ and  the $j$-th label of the sample $i$ individually. $\mathcal{D}=[\mathbf{d}_1;\mathbf{d}_2;\dots;\mathbf{d}_n]$ represents the label distribution of a batch. $\mathbf{d}_i=
[d_i^1,d_i^2,\dots, d_i^c]$ denotes the label distribution of the sample $i$, where $c$ and $d_i^j$ denotes the size of labels of the sample $i$ and the $j$-th label o
f the sample $i$ individually. Besides, $\mathbf{d}_i$ satisfies $\sum_{j=1}{d_i^j}=1$ and $0\le{d_i^j}\le{1}$.
\paragraph{Main Framework}
We follow the MulSupCon~\citep{Zhang2024MulSupCon} to use the MoCo~\citep{He2020MoCo} framework. We use $z_i^q$ and $z_i^k$ denote the $L_2$-normalized output of the query model and the key model individually for a sample $i$. The key model updates through a momentum-based method. In addition to utilizing a queue $\mathcal{Q}$ to store $z_i^k$, we also employ two queues $\mathcal{Q}_{ll}$ and $\mathcal{Q}_{ld}$ to store the logical labels $\mathbf{y}_i$ and the label distributions $\mathbf{d}_i$ of sample $i$, respectively.
\subsection{Multi-Label Supervised Contrastive Loss}
MulSupCon~\citep{Zhang2024MulSupCon} gives two preliminary ideas to define positive samples: \begin{enumerate*}[label=\arabic*)] 
    \item \textbf{ALL}: Samples whose logical labels completely match those of sample $i$ qualify as positive samples. 
    \item \textbf{ANY}: Samples that contain any of the labels in the logical labels of sample $i$ qualify as positive samples.
\end{enumerate*} Clearly, the mathematical formula for the set of positive samples $\mathcal{P}(i)$ of sample $i$ for \textbf{ALL} is:
\begin{align}
\mathcal{P}(i)=\left\{p|p\in{\mathcal{A}(i)}, \mathbf{y}_p=\mathbf{y}_i\right\}\ ,
\end{align}
and for \textbf{ANY} is:
\begin{align}
\mathcal{P}(i)=\left\{p|p\in{\mathcal{A}(i)}, \mathbf{y}_p\cap{\mathbf{y}_i}\ne\mathbf{\varnothing}\right\}\ ,
\end{align}
where $\mathcal{A}(i)$ denotes indices of all samples including $\mathcal{B}$ and $\mathcal{Q}$.
Accordingly, MulSupCon~\citep{Zhang2024MulSupCon} balance the loss weights between labels based on the \textbf{ALL} and \textbf{ANY} strategies, and the loss function for each anchor $i$ is:
\begin{align}
\mathcal{L}_i=\sum_{y_i^j\in\mathbf{y}_i}{\frac{-1}{|\mathcal{P}^j(i)|}\sum_{p\in{\mathcal{P}^j(i)}}{\log{\frac{\exp{(z_i\cdot{z_p}/\tau)}}{\sum_{a\in{\mathcal{A}(i)}}{\exp(z_i\cdot{z_a}/\tau)}}}}}\ ,
\end{align}
where $\tau$ is a temperature parameter~\citep{Chen2020SimCLR}. So the loss function of one batch is:
\begin{align}
\mathcal{L}=\frac{1}{\sum_i{|\mathbf{y}_i|}}\sum_i{\mathcal{L}_i}\ .
\end{align}

\subsection{\texorpdfstring{MulSupCon$_{\text{LD}}$}{}}
Multi-Label contrastive learning, such as MulSupCon~\citep{Zhang2024MulSupCon}, logical labels are directly incorporated into the model's training process. One advantage of this approach is that logical labels are typically straightforward to obtain, as multi-label datasets often provide them. However, logical labels can only indicate whether a label is relevant to an instance or not, and they fail to convey the relative importance among labels. Consequently, using logical labels to balance class loss makes it challenging to effectively enhance the model's performance. Therefore, we consider introducing label distributions into contrastive learning and leveraging these distributions to balance the loss across different classes. 

However, label distributions are not available in the datasets we use nor in those of the comparative methods, we need to recover the corresponding label distributions from the logical labels. Inspired by DLDL~\citep{Jia2023LabelDL}, we propose two contrastive learning-based methods for recovering label distributions from logical labels: one uses RBF and the other directly utilizing the contrastive loss function. We evaluate the performance of both methods in the experimental Section~\ref{experiments}. Since both of our methods build on contrastive learning, we first discuss the loss function of our contrastive learning approach and define positive samples. We use \textbf{ALL} to define positive samples and the loss of the query $i$ is:
\begin{align}
\mathcal{L}_{con,i}=\sum_{p\in{\mathcal{P}(i)}}{\log{\frac{\exp{(\mathrm{sim}(z_i\cdot{z_p})/\tau)}}{\sum_{a\in{\mathcal{A}(i)}}{\exp{(\mathrm{sim}(z_i\cdot{z_a})/\tau)}}}}} \ ,
\end{align}
where $\mathcal{P}(i)$ is the set of positive samples, $\mathcal{A}(i)$ includes samples from the current batch $\mathcal{B}$ and samples from the queue $\mathcal{Q}$, $\mathrm{sim}$ denotes cosine similarity and $\tau$ is a temperature parameter~\citep{Chen2020SimCLR}. Unlike DLDL~\citep{Jia2023LabelDL}, which uses the k-nearest neighbors of sample $i$ in RBF as similar samples, after defining our contrastive loss function, we can directly utilize the positive sample set $\mathcal{P}$ and employ the encoded vectors $z$ from the contrastive loss function within the RBF. Therefore, our loss function of RBF is: 
\begin{align}
\mathcal{C}_{i,p}^{(1)} &= \exp{\left(-||z_i-z_p||_2^2/{2\sigma^2}\right)} \ ,
\end{align}
where $\sigma$ is the hyper-parameter of the RBF kernel, and $p$ comes from the positive sample set $\mathcal{P}(i)$ of sample $i$. We set $\sigma$ to 0.01, following DLDL~\citep{Jia2023LabelDL}. Contrastive learning brings the features of samples similar to the anchor closer together in the feature space, while pushing apart those that are dissimilar. Consequently, the label distributions of similar samples are also expected to be similar. Then, we can establish constraints for both:
\begin{align}
\mathcal{G}_{i}^{(1)}=\sum_{p\in{\mathcal{P}(i)}} \mathcal{C}_{i,p}^{(1)}||d_i-d_p||_2^2 \ .
\end{align}

To reduce the influence of the model assigning a certain degree of description to labels that are irrelevant to the instance, we have imposed constraints on the label distribution. The formula for this constraint is as follows:
\begin{align}
\mathcal{H}_i=||\mathcal{Q}_{ll}^p-\mathcal{Q}_{ld}^p||_2^2 \ ,
\end{align}
where $\mathcal{Q}_{ll}^p$ and $\mathcal{Q}_{ld}^p$ denote the queues of logical labels and label distributions, respectively, formed by the positive sample pairs corresponding to the current sample $i$. Combining the above formulas, one of our final loss function for the label distribution takes the form:
\begin{align}
\mathcal{L}_{ld,i}^{(1)} &= \mathcal{G}_i^{(1)} + \alpha\mathcal{H}_i + \beta||\mathbf{W}||_F^2 \ ,
\end{align}
where $\alpha$ and $\beta$ are hpyer-parameters, and $\mathbf{W}$ is a weight matrix, which serves as the parameters of the final fully connected layer that generates the distribution of labels, and we use its Frobenius norm to control the complexity of the model. Inspired by DLDL~\citep{Jia2023LabelDL}, we directly establish the relationship between features and label distributions from the perspective of the contrastive loss function and propose another method for recovering label distributions. In this context, we only modify the definition of $\mathcal{C}_{i,p}^{(1)}$ as follows:
\begin{align}
\label{eq:per_constrastive}
\mathcal{C}_{i,p}^{(2)} &= \log{\frac{\exp{(\mathrm{sim}(z_i\cdot{z_p})/\tau)}}{\sum_{a\in{\mathcal{A}(i)}}{\exp{(\mathrm{sim}(z_i\cdot{z_a})/\tau)}}}} \ .
\end{align}

Finally, our other label distribution loss function takes the following form:
\begin{align}
\mathcal{L}_{ld,i}^{(2)} &= \mathcal{G}_i^{(2)} + \alpha\mathcal{H}_i + \beta||\mathbf{W}||_F^2 \ .
\end{align}

Once we obtain the label distribution, it becomes a natural course of action to utilize it to balance the loss among the labels, leading to the final loss function of query $i$:
\begin{align}
\label{eq:final-loss1}
\mathcal{L}_i^{(1)}&=-\sum_{y_{i}^{j}\in \mathbf{y}_i}{{\sum_{p\in \mathcal{P} ^j(i)}{d_{p}^{j}\cdot y_{p}^{j}\mathcal{C}_{i,p}^{(2)}}}} + \mathcal{L} _{ld,i}^{(1)}\ ,
\end{align}
and
\begin{align}
\label{eq:final-loss2}
\mathcal{L}_i^{(2)}&=-\sum_{y_{i}^{j}\in \mathbf{y}_i}{\sum_{p\in \mathcal{P} ^j(i)}{d_{p}^{j}\cdot y_{p}^{j}\mathcal{C}_{i,p}^{(2)}}} + \mathcal{L} _{ld,i}^{(2)} \ .
\end{align}

So, the loss function for a batch is:
\begin{align}
\label{eq:batch-final-loss1}
\mathcal{L}^{(1)}=\sum_{i\in{n}}{\mathcal{L}_i^{(1)}}\ ,
\end{align}
and 
\begin{align}
\label{eq:batch-final-loss2}
\mathcal{L}^{(2)}=\sum_{i\in{n}}{\mathcal{L}_i^{(2)}} \ .
\end{align}

We name these two methods of implementing \textbf{MulSupCon$_{\text{LD}}$}, which Equation~\eqref{eq:batch-final-loss1} and Equation~\eqref{eq:batch-final-loss2} describe, \textbf{MulSupCon$_{\text{RLD}}$} and \textbf{MulSupCon$_{\text{CLD}}$}, respectively.

\begin{algorithm}[tb]
\caption{MulSupCon$_{\text{LD}}$}
\begin{algorithmic}[1]
\label{alg:algorithm}
\REQUIRE A batch of data $\mathcal{B}$.
\PARAMETER $\alpha, \beta$ and $\sigma$.
\ENSURE loss for a batch and label distribution
\STATE \textcolor{gray}{\# aug means data augmentation}
\STATE $x_0, x_1 \gets \text{aug}(\mathcal{B}), \text{aug}(\mathcal{B})
$
\STATE \textcolor{gray}{\# encoder\_q and encoder\_k are the encoders in MoCo}
\STATE $z \gets \text{normalize}(\text{encoder\_q}(x_0))$
\STATE $k \gets \text{normalize}(\text{encoder\_k}(x_1))$ \textcolor{gray}{\# no gradient}
\STATE \textcolor{gray}{\# the label distribution for data $\mathcal{B}$ and fc means fully connected layer}
\STATE $\mathcal{D} \gets \text{softmax}(\text{fc}(q))$
\STATE $loss \gets 0$
\FOR{$i \gets 0$ \TO $z.\text{shape}[0]$}
    \STATE \textcolor{gray}{\# positive sample mask of sample $i$}
    \STATE $mask \gets y[i] \cap \mathcal{Q}_{ll}$
    \STATE \textcolor{gray}{\# encoders for positive sample of sample $i$}
    \STATE $z_p \gets \text{stack}(k[i], \mathcal{Q}[mask])$
    \STATE \textcolor{gray}{\# label distributions for positive sample of sample $i$}
    \STATE $d_p \gets \text{stack}(\mathcal{D}[i], \mathcal{Q}_{ld}[mask])$
    \STATE \textcolor{gray}{\# logical labels for positive sample}
    \STATE $y_p \gets \text{stack}(y[i], \mathcal{Q}_{ll}[mask])$
    \STATE \textcolor{gray}{\# we can get MulSupCon$_{\text{RLD}}$ method}
    \STATE $loss \gets$ calculate(Equation~\eqref{eq:final-loss1})
    \STATE \textcolor{gray}{\# or we can get MulSupCon$_{\text{CLD}}$ method}
    \STATE $loss \gets$ calculate(Equation~\eqref{eq:final-loss2})
\ENDFOR
\RETURN $loss, \mathcal{D}$
\end{algorithmic}
\end{algorithm}

\section{Experiments}\label{experiments}
Initially, we will utilize MulSupCon$_{\text{LD}}$ to pretrain models on various datasets. Subsequently, we will employ the fully connected layer, which generates label distributions from the pretrained models, along with the Binary Cross Entropy (BCE) Loss to further train on the training set, thereby adapting the models for multi-label classification tasks.
\paragraph{Experimental Datasets}
We will evaluate the performance of our model using image datasets commonly employed in multi-label learning, including MS-COCO~\citep{Tsung2014COCO}, NUS-WIDE~\citep{Chua2009NUSWIDE}, MIRFLICKR~\citep{Huiskes2008MIRFLICKR}, and PASCAL-VOC~\citep{Mark2010PASCAL}. Note that during training, we use sample data from the 21 most frequent categories in the NUS-WIDE dataset. In addition, to ensure the diversity of data types, we will also perform evaluations on the multi-label vector dataset sourced from Mulan.\footnote{http://mulan.sourceforge.net/datasets-mlc.html} The specific dataset information can be found in Table~\ref{tab:dataset}. 

\begin{table}
    \centering
    \begin{tabular}{cccc}
        \toprule
        Dataset  & \#Samples & \#Labels & \ Mean L/S \\
        \midrule
        MS-COCO     & 82081/40137          & 80    &  2.93/2.90  \\
        NUS-WIDE    & 150000/59347          & 81  &   2.41/2.40     \\
        MIRFLICKR  & 19664/4917          & 24      &  3.78/3.76 \\
        PASCAL & 5011/4952          & 20     &   1.46/1.42 \\
        \midrule
        \midrule
        Bookmarks & 60000/27856 & 208 & 2.03/2.03 \\
        Mediamill & 29804/12373 & 101 & 4.54/4.60 \\
        Delicious & 12886/3181 & 983 & 19.07/18.94 \\
        Scene & 1210/1195 & 6 & 1.06/1.09 \\
        Yeast & 1500/917 & 14 & 4.23/4.25 \\
        \bottomrule
    \end{tabular}
    \caption{Details of the image and vector datasets. Mean L/S represents the average number of labels per sample in the dataset and use a slash to divide the information of the training and test datasets.}
    \label{tab:dataset}
\end{table}

\paragraph{Metrics}
We follow MulSupCon~\citep{Zhang2024MulSupCon} to employ six widely-used metrics to evaluate the performance of our method as well as the comparative methods. The six metrics are:  \begin{enumerate*}[label=\arabic*)] 
    \item the mean value of average (mAP), which provides a comprehensive global perspective for evaluating a model's performance across all labels by calculating the average proportion of correct predictions, ranked by confidence, within the prediction results for each label. 
    \item the precision of the top-1 (precision@1), which measures whether the label predicted by the model as the most probable for each instance is correct, 
    \item macro-F1, which assigns equal weight to each label, irrespective of the label distribution, 
    \item micro-F1, takes into account the distribution of labels, where errors in common labels have a greater impact on the overall performance, 
    \item Hamming Accuracy $\frac{1}{c}\sum_{j=1}^c{\mathds{1}[y_i^j=\hat{y}_i^j]}\ ,$ which directly reflects the degree of correspondence between the predicted labels and the true labels, and 
    \item example-based F1 $\frac{2\sum_{i=1}^c{y_i\hat{y}_i}}{\sum_{i=1}^c{y_i}+\sum_{i=1}^c{\hat{y}_i}}\ ,$ where $\hat{y}$ is the predicted label and $c$ is the number of class.
\end{enumerate*}
Next, we will use mAP, p\_at\_1, maF1, miF1, HA and ebF1 to represent these metrics respectively.

\paragraph{Settings}
We use ResNet-50 as a common encoder for \textbf{image} \textbf{datasets}. Throughout the training phase, we will resize the images to 224 $\times$ 224 and employ the data augmentation scheme from SimCLR~\citep{Chen2020SimCLR}. During the pretraining phase, we configure the number of epochs to 400 and utilize AdamW and CosineAnnealingWarmRestarts as the optimizer and scheduler. The batch size for all image datasets is 128. For MS-COCO dataset, NUS-WIDE dataset, MIRFLICKR dataset and PASCAL dataset, during the pretraining phase, we use a momentum of 0.999, a temperature parameter of 0.1, a queue size of 4096, except MIRFLICKR is 8192 and PASCAL is 1024, a learning rate of 0.00004, and a weight decay of 0.0001. The parameters in the learning rate scheduler are set to 50, 2, and 0.00001, respectively. After the pretraining phase, the learning rate is set to 0.0004, the weight decay to 0.00001, and the learning rate scheduler parameters to 25, 2, and 0.0001, respectively.

For \textbf{vector} \textbf{datasets}, we employ a straightforward multilayer perceptron composed of three fully connected layers, ReLU activation functions, and Dropout. During the pretraining phase, we randomly mask 50\% of the elements. We use the same optimizer and scheduler as those in image datasets. We set the number of epochs to 400. The batch size is 256 for Bookmarks and Scene, 128 for Delicious and Yeast, and 512 for Mediamill. For Yeast, Scene, Bookmarks, Delicious and Mediamill dataset, during the pretraining phase, we randomly mask 0.5 of the elements. we use a momentum of 0.999, a temperature parameter of 0.1, a queue size of 8192 except Yeast is 256, Scene is 1024 and Delicious is 4096, a learning rate of 0.00004, and a weight decay of 0.0001. The parameters in the learning rate scheduler are set to 50, 2, and 0.00001, respectively. After the pretraining phase, the learning rate is set to 0.0004, weight decay to 0.00001, except Mediamill is 0.004 and 0.0001. All masks are set to 0.4, except for Delicious and Mediamill, which are set to 0.2 and 0. The parameters of Scene and Bookmarks in the learning rate scheduler are set to 25, 2, and 0.0001. The parameters of Scene and Delicious in the learning rate scheduler are set to 50, 2, and 0.0001. The parameters of Mediamill in the learning rate scheduler are set to 50, 2, and 0.001.

After pretraining, we train the model using BCE with the number of epochs set to 100 for all datasets.

\paragraph{Comparison Methods}
Our method will be compared with the following methods: 
\begin{enumerate}[label=\arabic*)] 
    \item \textbf{LaMP}~\citep{Lanchantin2020LaMP}: LaMP model is to treat labels as nodes in a graph and perform neural message passing through an attention mechanism, thereby effectively modeling the joint prediction of multiple labels. 
    \item \textbf{MPVAE}~\citep{Bai2022MPVAE}: MPVAE combines a variational autoencoder with a multivariate normal distribution model, effectively learning the label embedding space and the correlations between labels.
    \item \textbf{ASL}~\citep{Ridnik2021ASL}: ASL introduces a loss function called asymmetric loss, which addresses the issue of imbalance between positive and negative samples in multi-label classification problems.
    \item \textbf{RBCC}~\citep{Gerych2021RBCC}: RBCC improves prediction accuracy by using a Bayesian network that learns class dependency relationships.
    \item \textbf{C-GMVAE}~\citep{Bai2022gaussian}: C-GMVAE model map features and labels into a shared latent space and capture the relationships between features and labels through contrastive learning.
    \item \textbf{MulSupCon}~\citep{Zeng2024Multi-Label}: Building on the \textbf{ALL} and \textbf{ANY} method, this method separately considers the labels of samples to balance the loss weights among different labels.
\end{enumerate}

\subsection{Performance Comparison}
In this section, we compare the performance differences between our two proposed methods,  MulSupCon$_{\text{RLD}}$ and MulSupCon$_{\text{CLD}}$, for implementing MulSupCon$_{\text{LD}}$ and other multi-label classification methods on both image datasets and vector datasets. The evaluation results for the vector datasets partially come from C-GMVAE, except for MulSupCon, whose results come from our replication on the same datasets. We also evaluate MulSupCon on image datasets, and we can find these results from Table~\ref{tab:metrics-comparison-of-all-methods-1} to Table~\ref{tab:metrics-comparison-of-all-methods-3}. From the experimental results, our methods demonstrate superior performance on both vector and image datasets. 

We believe the reasons for outperforming other comparative methods are twofold. On one hand, we use label distributions to balance the loss across different classes. On the other hand, we incorporate sufficiently effective supervisory information during the recovery of label distributions, enabling the model to assign description degrees to labels that are most relevant to the instances as much as possible.
However, overall, MulSupCon$_{\text{CLD}}$ outperforms MulSupCon$_{\text{RLD}}$. We believe the reason is that MulSupCon$_{\text{RLD}}$ not only introduces additional computational overhead but, more importantly, struggles to effectively recover label distributions from logical labels. 

\begin{table*}[htbp]
    \centering
    \resizebox{\linewidth}{!}{
    \begin{tabular}{cccccc|ccccc}
    \toprule
    Algorithm & Yeast & Scene & Bookmarks & Delicious & Mediamill & Yeast & Scene & Bookmarks & Delicious & Mediamill \\
    \midrule
    HA $\uparrow$ &  &  &  &  &  & ~ebF1 $\uparrow$ &  &  &  &   \\
    \midrule
    LaMP & 0.786 & 0.903 & \textbf{0.992} & 0.982 & - & 0.624 & 0.728 & 0.389 & 0.372 & - \\
    MPVAE & 0.792 & 0.909 & 0.991 & 0.982 & - & 0.648 & 0.751 & 0.382 & 0.373 & - \\
    ASL & 0.796 & 0.912 & 0.991 & 0.982 & - & 0.613 & 0.770 & 0.373 & 0.359 & - \\
    RBCC & 0.793 & 0.904 & - & - & - & 0.605 & 0.758 & - & - & - \\
    C-GMVAE & 0.796 & 0.915 & \textbf{0.992} & \textbf{0.983} & 0.970 & 0.656 & 0.777 & 0.392 & 0.381 & \textbf{0.623} \\
    MulSupCon & 0.803 & 0.925 & \textbf{0.992} & 0.982 & 0.970 & 0.658 & 0.792 & 0.400 & 0.374 & 0.569 \\
    MulSupCon$_\text{RLD}$ & 0.802 & \textbf{0.930} & 0.991 & 0.982 & \textbf{0.971} & 0.661 & \textbf{0.815} & \textbf{0.405} & 0.382 & 0.597 \\
    MulSupCon$_\text{CLD}$ & \textbf{0.805} & 0.929 & 0.991 & 0.982 & 0.970 & \textbf{0.664} & 0.813 & 0.400 & \textbf{0.384} & 0.620 \\
    \midrule
    miF1 $\uparrow$ &  &  &  &  &  & ~maF1 $\uparrow$ &  &  &  &   \\
    \midrule
    LaMP & 0.641 & 0.716 & 0.373 & 0.386 & - & 0.480 & 0.745 & 0.286 & 0.196 & - \\
    MPVAE & 0.655 & 0.742 & 0.375 & 0.393 & - & 0.482 & 0.750 & 0.285 & 0.181 & - \\
    ASL & 0.637 & 0.753 & 0.354 & 0.387 & - & 0.484 & 0.765 & 0.264 & 0.183 & - \\
    RBCC & 0.623 & 0.749 & - & - & - & 0.480 & 0.753 & - & - & - \\
    C-GMVAE & 0.665 & 0.762 & 0.377 & \textbf{0.403} & \textbf{0.626} & 0.487 & 0.769 & 0.291 & \textbf{0.197} & \textbf{0.273} \\
    MulSupCon & 0.668 & 0.782 & 0.402 & 0.391 & 0.589 & 0.464 & \textbf{0.787} & 0.307 & 0.153 & 0.146 \\
    MulSupCon$_\text{RLD}$ & 0.671 & \textbf{0.807} & \textbf{0.405} & 0.398 & 0.618 & 0.486 & 0.770 & 0.311 & 0.188 & 0.262 \\
    MulSupCon$_\text{CLD}$ & \textbf{0.676} & 0.805 & 0.403 & 0.401 & 0.624 & \textbf{0.496} & 0.766 & \textbf{0.318} & 0.195 & 0.262 \\
    \bottomrule
    \end{tabular}
    }
    \caption{The results of the compared methods in terms of four metrics on multi-label vector datasets and highlights the best results in boldface. For all four metrics, higher values indicate better performance and use the symbol $\uparrow$ to illustrate it. The symbol - means that the original paper does not provide values for this metric.}
    \label{tab:metrics-comparison-of-all-methods-1}
\end{table*}

\begin{table*}[htbp]
    \centering
    \resizebox{\linewidth}{!}{
    \begin{tabular}{cccccc|ccccc}
    \toprule
    Algorithm & Yeast & Scene & Bookmarks & Delicious & Mediamill & Yeast & Scene & Bookmarks & Delicious & Mediamill \\
    \midrule
    p\_at\_1 $\uparrow$ &  &  &  &  &  & ~mAP $\uparrow$ &  &  &  &   \\
    \midrule
    MulSupCon & 0.761 & 0.811 & \textbf{0.496} & 0.681 & 0.837 & 0.480 & \textbf{0.859} & 0.264 & 0.155 & 0.150 \\
    MulSupCon$_\text{RLD}$ & 0.762 & \textbf{0.832} & 0.492 & \textbf{0.691} & 0.852 & 0.514 & 0.825 & 0.273 & 0.181 & 0.257 \\
    MulSupCon$_\text{CLD}$ & \textbf{0.769} & 0.826 & 0.491 & 0.685 & \textbf{0.889} & \textbf{0.521} & 0.821 & \textbf{0.275} & \textbf{0.185} & \textbf{0.261} \\
    \bottomrule
    \end{tabular}
    }
    \caption{An extension to Table~\ref{tab:metrics-comparison-of-all-methods-1}. The evaluation results of methods MulSupCon, MulSupCon$_\text{RLD}$, and MulSupCon$_\text{CLD}$ on the p\_at\_1 and mAP metrics, as other methods lack evaluation values for these two metrics.}
    \label{tab:metrics-comparison-of-all-methods-2}
\end{table*}

\begin{table*}[htbp]
    \centering
    \resizebox{\linewidth}{!}{
    \begin{tabular}{ccccc|cccc}
    \toprule
    Algorithm & MIRFLICKR & NUS-WIDE & MS-COCO & PASCAL & MIRFLICKR & NUS-WIDE & MS-COCO & PASCAL \\
    \midrule
    HA $\uparrow$ &  &  &  &    & ~ebF1 $\uparrow$ &  &   &   \\
    \midrule
    MulSupCon & 0.9130 & \textbf{0.9275} & 0.9769 & 0.9485 & 0.6729 & 0.7449 & 0.6334 & 0.6004 \\
    MulSupCon$_\text{RLD}$ & \textbf{0.9144} & 0.9257 & 0.9747 & 0.9495 & \textbf{0.6759} & 0.7407 & 0.5770 & 0.6043 \\
    MulSupCon$_\text{CLD}$ & 0.9131 & \textbf{0.9275} & \textbf{0.9780} & \textbf{0.9499} & 0.6728 & \textbf{0.7485} & \textbf{0.6565} & \textbf{0.6104} \\
    \midrule
    miF1 $\uparrow$ &  &    &  &  & ~maF1 $\uparrow$ &  &   &   \\
    \midrule
    MulSupCon & 0.7011 & 0.7731 & 0.6245 & 0.6136 & \textbf{0.6074} & 0.7259 & 0.5513 & 0.5296 \\
    MulSupCon$_\text{RLD}$ & \textbf{0.7076} & 0.7694 & 0.5865 & 0.6180 & 0.6050 & 0.7203 & 0.4974 & 0.5513 \\
    MulSupCon$_\text{CLD}$ & 0.7023 & \textbf{0.7744} & \textbf{0.6481} & \textbf{0.6234} & 0.6049 & \textbf{0.7265} & \textbf{0.5790} & \textbf{0.5535} \\
    \midrule
    p\_at\_1 $\uparrow$ &   &  &  &  & ~mAP $\uparrow$ &  &  &   \\
    \midrule
    MulSupCon & 0.8514 & 0.8618 & 0.8562 & 0.6976 & 0.6579 & 0.7859 & 0.5756 & 0.5843 \\
    MulSupCon$_\text{RLD}$ & 0.8598 & 0.8599 & 0.7996 & 0.7138 & \textbf{0.6615} & 0.7804 & 0.5149 & 0.5909 \\
    MulSupCon$_\text{CLD}$ & \textbf{0.8633} & \textbf{0.8711} & \textbf{0.8769} & \textbf{0.7142} & 0.6600 & \textbf{0.7868} & \textbf{0.6113} & \textbf{0.5950} \\
    \bottomrule
    \end{tabular}
    }
    \caption{The results of methods MulSupCon, MulSupCon$_\text{RLD}$, and MulSupCon$_\text{CLD}$ on six metrics for image datasets and highlight the best results in boldface. For all six metrics, higher values indicate better performance and use the symbol $\uparrow$ to illustrate the metrics.}
    \label{tab:metrics-comparison-of-all-methods-3}
\end{table*}

\begin{figure*}[htbp]
    \centering
    \begin{subfigure}[ht]{\textwidth}
        \vspace{0.8cm}
        \centering
        \includegraphics[width=0.7\textwidth]{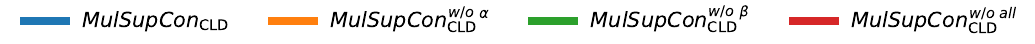}
        \label{fig:ablation-study-legend}
    \end{subfigure}
    \vfill
    \begin{subfigure}[ht]{\textwidth}
        \includegraphics[width=0.15\textwidth]{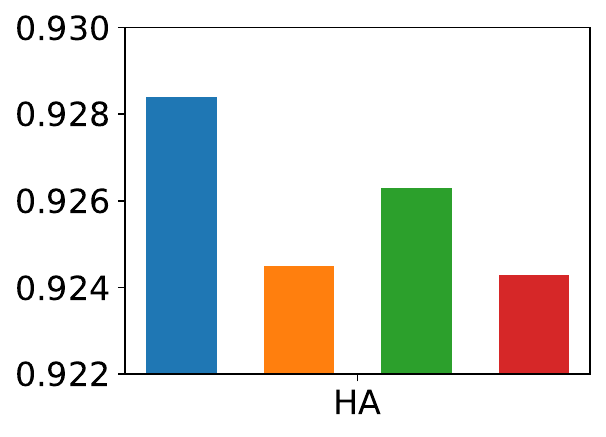}\hfill
        \includegraphics[width=0.15\textwidth]{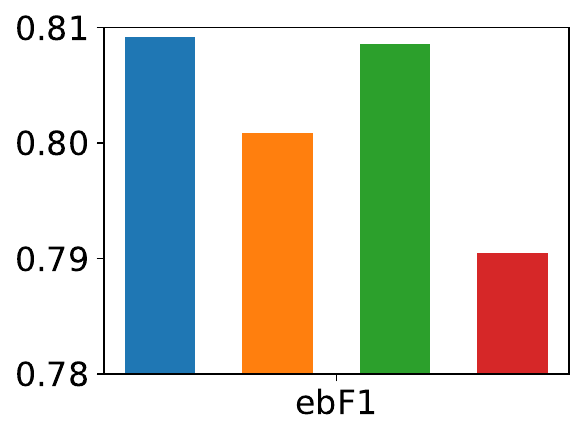}\hfill
        \includegraphics[width=0.15\textwidth]{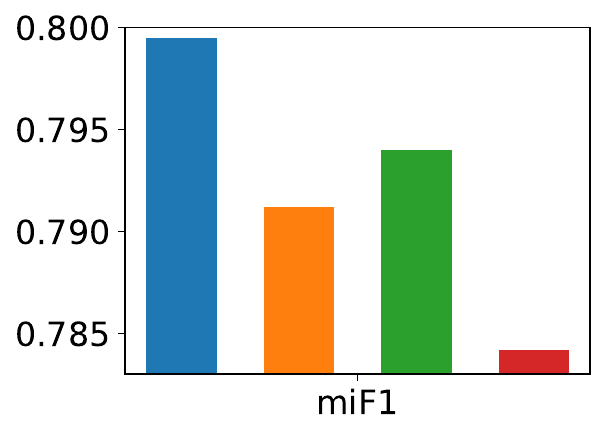}\hfill
        \includegraphics[width=0.15\textwidth]{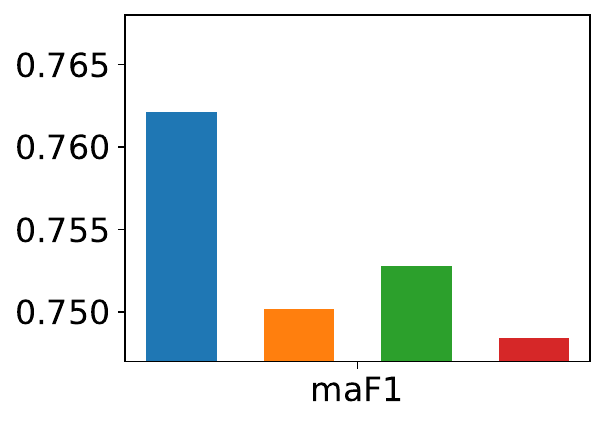}\hfill
        \includegraphics[width=0.15\textwidth]{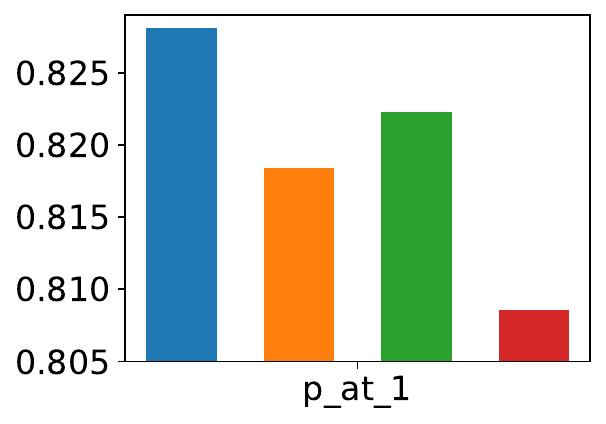}\hfill
        \includegraphics[width=0.15\textwidth]{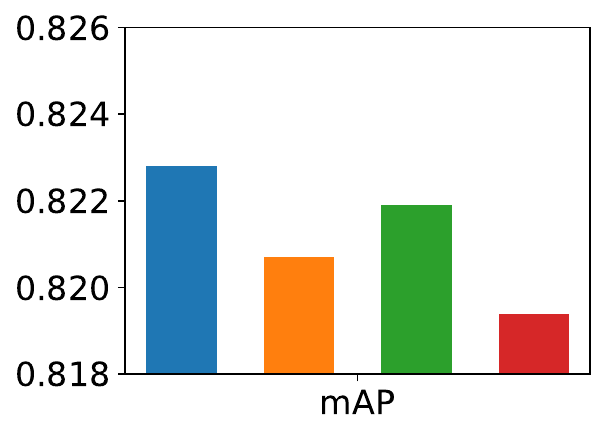}
        \caption{Scene dataset.}
        \label{fig:ablation-study-1}
    \end{subfigure}
    \vfill
    \begin{subfigure}[ht]{\textwidth}
        \includegraphics[width=0.15\textwidth]{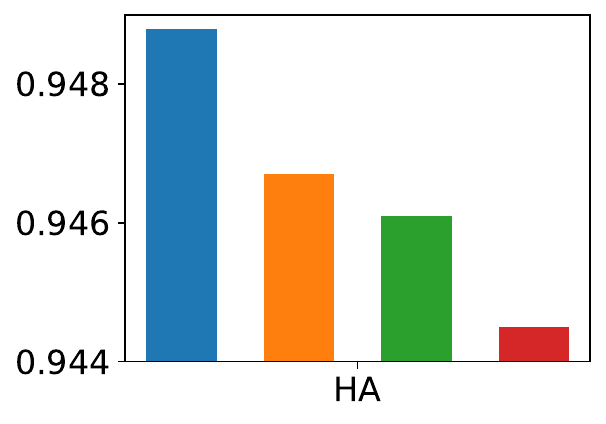}\hfill
        \includegraphics[width=0.15\textwidth]{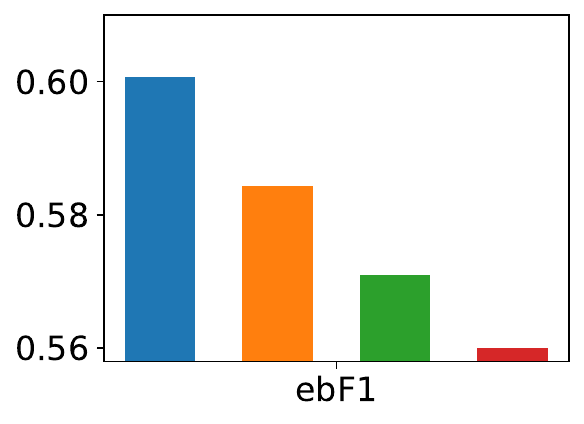}\hfill
        \includegraphics[width=0.15\textwidth]{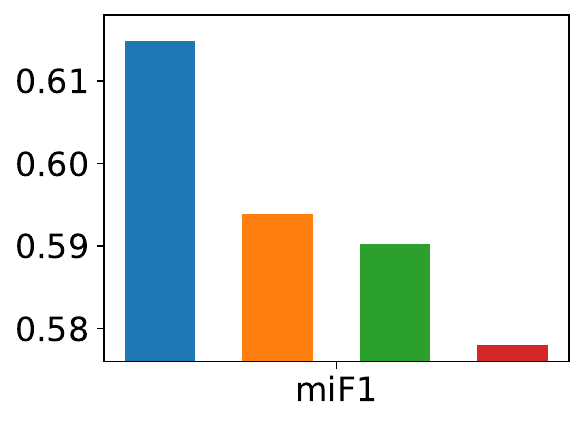}\hfill
        \includegraphics[width=0.15\textwidth]{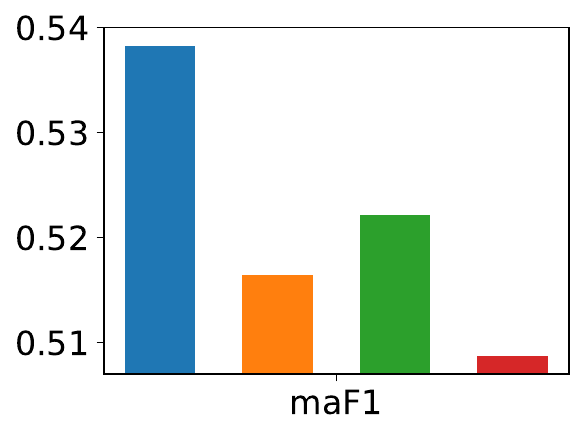}\hfill
        \includegraphics[width=0.15\textwidth]{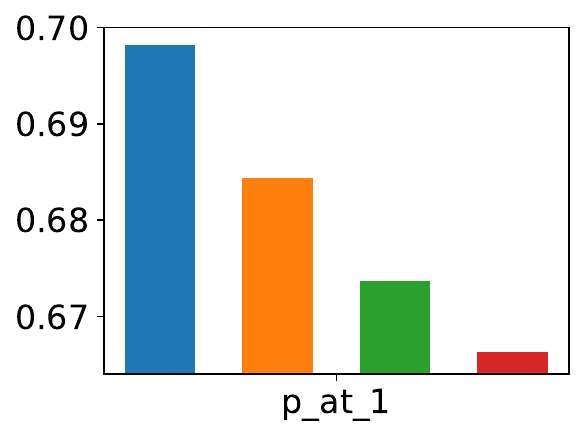}\hfill
        \includegraphics[width=0.15\textwidth]{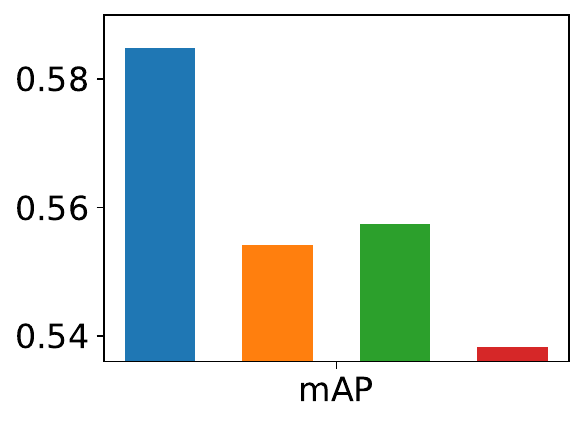}
        \caption{PASCAL dataset.
        }
        \label{fig:ablation-study-2}
    \end{subfigure}
    \caption{Ablation studies on two datasets across six metrics validate the effectiveness of label distribution recovery modules ($\alpha$ and $\beta$).}
    \label{fig:ablation-study}
\end{figure*}

\begin{table*}[ht]
    \centering

        \begin{subtable}{\textwidth}
        \centering
        \begin{tabular}{lcccccc}
            \toprule
            $\beta$ & HA & ebF1 & miF1 & maF1 & p\_at\_1 & mAP \\
            \midrule
            0.001 & 0.925 & 0.800 & 0.790 & 0.749 & 0.821 & 0.817 \\
            0.01 & 0.924 & 0.804 & 0.792 & 0.751 & 0.815 & 0.817 \\
            0.1 & 0.925 & 0.801 & 0.794 & 0.753 & 0.818 & \textbf{0.820} \\
            1 & \textbf{0.928} & \textbf{0.813} & \textbf{0.804} & \textbf{0.765} & \textbf{0.823} & 0.819 \\
            \bottomrule
        \end{tabular}
        \caption{Fix the parameter $\alpha$ at 0.001 on the Scene dataset.}
        \label{tab:psa-scene-1}
    \end{subtable}
    \vfill
    \begin{subtable}{\textwidth}
        \centering
        \begin{tabular}{lcccccc}
            \toprule
            $\beta$ & HA & ebF1 & miF1 & maF1 & p\_at\_1 & mAP \\
            \midrule
            0.001 & 0.927 & 0.809 & 0.795 & 0.754 & \textbf{0.825} & \textbf{0.822} \\
            0.01 & 0.925 & 0.800 & 0.792 & 0.750 & 0.819 & 0.820 \\
            0.1 & 0.924 & 0.798 & 0.793 & 0.752 & 0.819 & 0.817 \\
            1 & \textbf{0.928} & \textbf{0.813} & \textbf{0.803} & \textbf{0.763} & 0.819 & 0.821 \\
            \bottomrule
        \end{tabular}
    \caption{Fix the parameter $\alpha$ at 0.01 on the Scene dataset.}
    \label{tab:psa-scene-2}
    \end{subtable} 
    
    \begin{subtable}{\textwidth}
        \centering
        \begin{tabular}{lcccccc}
            \toprule
            $\beta$ & HA & ebF1 & miF1 & maF1 & p\_at\_1 & mAP \\
            \midrule
            0.001 & 0.926 & 0.806 & 0.796 & 0.753 & 0.823 & \textbf{0.823} \\
            0.01 & 0.925 & 0.800 & 0.794 & 0.753 & 0.820 & 0.818 \\
            0.1 & 0.925 & 0.800 & 0.790 & 0.749 & 0.821 & 0.817 \\
            1 & \textbf{0.929} & \textbf{0.813} & \textbf{0.805} & \textbf{0.766} & \textbf{0.826} & 0.821 \\
            \bottomrule
        \end{tabular}
        \caption{Fix the parameter $\alpha$ at 0.1 on the Scene dataset.}
        \label{tab:psa-scene-3}
    \end{subtable}
    \vfill
    \begin{subtable}{\textwidth}
        \centering
        \begin{tabular}{lcccccc}
            \toprule
            $\beta$ & HA & ebF1 & miF1 & maF1 & p\_at\_1 & mAP \\
            \midrule
            0.001 & 0.926 & 0.803 & 0.791 & 0.750 & 0.816 & 0.819 \\
            0.01 & 0.926 & 0.804 & 0.793 & 0.754 & \textbf{0.822} & \textbf{0.821} \\
            0.1 & 0.925 & 0.801 & 0.791 & 0.750 & 0.820 & 0.819 \\
            1 & \textbf{0.928} & \textbf{0.807} & \textbf{0.797} & \textbf{0.758} & 0.820 & 0.821 \\
            \bottomrule
        \end{tabular}
        \caption{Fix the parameter $\alpha$ at 1 on the Scene dataset.}
        \label{tab:psa-scene-4}
    \end{subtable}
    \caption{Parameter sensitivity experiments for the method MulSupCon$_{\text{CLD}}$ on the Scene and only fix $\alpha$}
    \label{tab:psa-1}
\end{table*}

\begin{table}[ht]
    \centering
     \begin{subtable}{\textwidth}
        \centering
        \begin{tabular}{lcccccc}
            \toprule
            $\alpha$ & HA & ebF1 & miF1 & maF1 & p\_at\_1 & mAP \\
            \midrule
            0.001 & 0.925 & 0.800 & 0.790 & 0.749 & 0.821 & 0.817 \\
            \midrule
            0.01 & \textbf{0.927} & \textbf{0.809} & \textbf{0.795} & \textbf{0.754} & \textbf{0.825} & \textbf{0.822} \\
            \midrule
            0.1 & 0.925 & 0.800 & 0.794 & 0.753 & 0.820 & 0.818 \\
            \midrule
            1 & 0.926 & 0.803 & 0.791 & 0.750 & 0.816 & 0.819 \\
            \bottomrule
        \end{tabular}
        \caption{Fix the parameter $\beta$ at 0.001 on the Scene dataset.}
        \label{tab:psa-scene-5}
        \end{subtable}
        \vfill
        \begin{subtable}{\textwidth}
        \centering
        \begin{tabular}{lcccccc}
            \toprule
            $\alpha$ & HA & ebF1 & miF1 & maF1 & p\_at\_1 & mAP \\
            \midrule
            0.001 & 0.924 & \textbf{0.804} & 0.792 & 0.751 & 0.815 & 0.817 \\
            \midrule
            0.01 & 0.925 & 0.800 & 0.792 & 0.750 & 0.819 & 0.820 \\
            \midrule
            0.1 & 0.925 & 0.800 & \textbf{0.794} & 0.753 & 0.820 & 0.818 \\
            \midrule
            1 & \textbf{0.926} & \textbf{0.804} & 0.793 & \textbf{0.754} & \textbf{0.822} & \textbf{0.821} \\
            \bottomrule
        \end{tabular}
        \caption{Fix the parameter $\beta$ at 0.01 on the Scene dataset.}
        \label{tab:psa-scene-6}
        \end{subtable}
    \vfill
    \begin{subtable}{\textwidth}
    \centering
    \begin{tabular}{lcccccc}
        \toprule
        $\alpha$ & HA & ebF1 & miF1 & maF1 & p\_at\_1 & mAP \\
        \midrule
        0.001 & \textbf{0.925} & \textbf{0.801} & \textbf{0.794} & \textbf{0.753} & 0.818 & \textbf{0.820} \\
        \midrule
        0.01 & 0.924 & 0.798 & 0.793 & 0.752 & 0.819 & 0.817 \\
        \midrule
        0.1 & \textbf{0.925} & 0.800 & 0.790 & 0.749 & \textbf{0.821} & 0.817 \\
        \midrule
        1 & \textbf{0.925} & \textbf{0.801} & 0.791 & 0.750 & 0.820 & 0.819 \\
        \bottomrule
    \end{tabular}
    \caption{Fix the parameter $\beta$ at 0.1 on the Scene dataset.}
    \label{tab:psa-scene-7}
    \end{subtable}
    \vfill
    \begin{subtable}{\textwidth}
    \centering
    \begin{tabular}{lcccccc}
        \toprule
        $\alpha$ & HA & ebF1 & miF1 & maF1 & p\_at\_1 & mAP \\
        \midrule
        0.001 & 0.928 & \textbf{0.813} & 0.804 & 0.765 & 0.823 & 0.819 \\
        \midrule
        0.01 & 0.928 & \textbf{0.813} & 0.803 & 0.763 & 0.819 & \textbf{0.821} \\
        \midrule
        0.1 & \textbf{0.929} & \textbf{0.813} & \textbf{0.805} & \textbf{0.766} & \textbf{0.826} & \textbf{0.821} \\
        \midrule
        1 & 0.928 & 0.807 & 0.797 & 0.758 & 0.820 & \textbf{0.821} \\
        \bottomrule
    \end{tabular}
    \caption{Fix the parameter $\beta$ at 1 on the Scene dataset.}
    \label{tab:psa-scene-8}
    \end{subtable}
    \caption{An extension to Table~\ref{tab:psa-1} and only fix $\beta$}
    \label{tab:psa-2}
\end{table}

\begin{table}[ht]
    \centering
    \begin{subtable}{\textwidth}
        \centering
        \begin{tabular}{lcccccc}
            \toprule
            $\beta$ & HA & ebF1 & miF1 & maF1 & p\_at\_1 & mAP \\
            \midrule
            0.001 & 0.946 & 0.575 & 0.596 & 0.518 & 0.671 & 0.557 \\
            0.01 & 0.9476 & 0.594 & 0.607 & 0.547 & 0.691 & 0.576 \\
            0.1 & 0.947 & 0.589 & 0.602 & 0.527 & 0.687 & 0.562 \\
            1 & \textbf{0.948} & \textbf{0.596} & \textbf{0.611} & \textbf{0.550} & \textbf{0.705} & \textbf{0.595} \\
            \bottomrule
        \end{tabular}
        \caption{Fix the parameter $\alpha$ at 0.001 on the PASCAL dataset.}
        \label{tab:psa-pascal-1}
    \end{subtable}
    \hfill
    \begin{subtable}{\textwidth}
        \centering
        \begin{tabular}{lcccccc}
            \toprule
            $\beta$ & HA & ebF1 & miF1 & maF1 & p\_at\_1 & mAP \\
            \midrule
            0.001 & 0.946 & 0.586 & 0.596 & 0.523 & 0.684 & 0.566 \\
            0.01 & \textbf{0.948} & \textbf{0.595} & \textbf{0.610} & \textbf{0.544} & \textbf{0.700} & \textbf{0.582} \\
            0.1 & 0.946 & 0.594 & 0.601 & 0.539 & 0.685 & 0.571 \\
            1 & 0.946 & 0.594 & 0.601 & 0.539 & 0.685 & 0.571 \\
            \bottomrule
        \end{tabular}
    \caption{Fix the parameter $\alpha$ at 0.01 on the PASCAL dataset.}
    \label{tab:psa-pascal-2}
    \end{subtable}
    \vfill
    \begin{subtable}{\textwidth}
        \centering
        \begin{tabular}{lcccccc}
            \toprule
            $\beta$ & HA & ebF1 & miF1 & maF1 & p\_at\_1 & mAP \\
            \midrule
            0.001 & \textbf{0.947} & \textbf{0.593} & \textbf{0.605} & 0.529 & \textbf{0.692} & \textbf{0.568} \\
            0.01 & 0.945 & 0.576 & 0.591 & 0.514 & 0.659 & 0.565 \\
            0.1 & 0.946 & 0.583 & 0.597 & \textbf{0.538} & 0.680 & 0.579 \\
            1 & 0.945 & 0.571 & 0.586 & 0.489 & 0.669 & 0.551 \\
            \bottomrule
        \end{tabular}
        \caption{Fix the parameter $\alpha$ at 0.1 on the PASCAL dataset.}
        \label{tab:psa-pascal-3}
    \end{subtable}
    \hfill
    \begin{subtable}{\textwidth}
        \centering
        \begin{tabular}{lcccccc}
            \toprule
            $\beta$ & HA & ebF1 & miF1 & maF1 & p\_at\_1 & mAP \\
            \midrule
            0.001 & \textbf{0.949} & \textbf{0.610} & \textbf{0.623} & \textbf{0.553} & \textbf{0.714} & \textbf{0.595} \\
            0.01 & 0.946 & 0.575 & 0.588 & 0.504 & 0.670 & 0.549 \\
            0.1 & 0.945 & 0.580 & 0.591 & 0.523 & 0.667 & 0.567 \\
            1 & 0.947 & 0.582 & 0.598 & 0.543 & 0.685 & 0.583 \\
            \bottomrule
        \end{tabular}
        \caption{Fix the parameter $\alpha$ at 1 on the PASCAL dataset.}
        \label{tab:psa-pascal-4}
    \end{subtable}
    \caption{Parameter sensitivity experiments for the method MulSupCon$_{\text{CLD}}$ on the PASCAL and only fix $\alpha$}
    \label{tab:psa-3}
\end{table}

\begin{table}[ht]
    \centering
    
        \begin{subtable}{\textwidth}
        \centering
        \begin{tabular}{lcccccc}
            \toprule
            $\alpha$ & HA & ebF1 & miF1 & maF1 & p\_at\_1 & mAP \\
            \midrule
            0.001 & 0.946 & 0.575 & 0.596 & 0.518 & 0.671 & 0.557 \\
            \midrule
            0.01 & 0.946 & 0.586 & 0.596 & 0.523 & 0.684 & 0.566 \\
            \midrule
            0.1 & 0.947 & 0.593 & 0.605 & 0.529 & 0.692 & 0.568 \\
            \midrule
            1 & \textbf{0.949} & \textbf{0.610} & \textbf{0.623} & \textbf{0.553} & \textbf{0.714} & \textbf{0.595} \\
            \bottomrule
        \end{tabular}
        \caption{Fix the parameter $\beta$ at 0.001 on the PASCAL dataset.}
        \label{tab:psa-pascal-5}
        \end{subtable}
        \hfill
        \begin{subtable}{\textwidth}
        \centering
        \begin{tabular}{lcccccc}
            \toprule
            $\alpha$ & HA & ebF1 & miF1 & maF1 & p\_at\_1 & mAP \\
            \midrule
            0.001 & 0.947 & 0.594 & 0.607 & \textbf{0.547} & 0.691 & 0.576 \\
            \midrule
            0.01 & \textbf{0.948} & \textbf{0.595} & \textbf{0.610} & 0.544 & \textbf{0.700} & \textbf{0.582} \\
            \midrule
            0.1 & 0.945 & 0.576 & 0.591 & 0.514 & 0.659 & 0.565 \\
            \midrule
            1 & 0.946 & 0.575 & 0.588 & 0.504 & 0.670 & 0.549 \\
            \bottomrule
        \end{tabular}
        \caption{Fix the parameter $\beta$ at 0.01 on the PASCAL dataset.}
        \label{tab:psa-pascal-6}
        \end{subtable}

    \vfill
    \begin{subtable}{\textwidth}
    \centering
    \begin{tabular}{lcccccc}
        \toprule
        $\alpha$ & HA & ebF1 & miF1 & maF1 & p\_at\_1 & mAP \\
        \midrule
        0.001 & \textbf{0.947} & 0.589 & \textbf{0.602} & 0.527 & \textbf{0.687} & 0.562 \\
        \midrule
        0.01 & 0.946 & \textbf{0.594} & 0.601 & \textbf{0.539} & 0.685 & 0.571 \\
        \midrule
        0.1 & 0.946 & 0.583 & 0.597 & 0.538 & 0.680 & \textbf{0.579} \\
        \midrule
        1 & 0.945 & 0.580 & 0.591 & 0.523 & 0.667 & 0.567 \\
        \bottomrule
    \end{tabular}
    \caption{Fix the parameter $\beta$ at 0.1 on the PASCAL dataset.}
    \label{tab:psa-pascal-7}
    \end{subtable}
    \hfill
    \begin{subtable}{\textwidth}
    \centering
    \begin{tabular}{lcccccc}
        \toprule
        $\alpha$ & HA & ebF1 & miF1 & maF1 & p\_at\_1 & mAP \\
        \midrule
        0.001 & \textbf{0.948} & \textbf{0.596} & \textbf{0.611} & \textbf{0.550} & \textbf{0.705} & \textbf{0.595} \\
        \midrule
        0.01 & 0.946 & 0.594 & 0.601 & 0.539 & 0.685 & 0.571 \\
        \midrule
        0.1 & 0.945 & 0.571 & 0.586 & 0.489 & 0.669 & 0.551 \\
        \midrule
        1 & 0.947 & 0.582 & 0.598 & 0.543 & 0.685 & 0.583 \\
        \bottomrule
    \end{tabular}
    \caption{Fix the parameter $\beta$ at 1 on the PASCAL dataset.}
    \label{tab:psa-pascal-8}
    \end{subtable}
    \caption{An extension to Table~\ref{tab:psa-3} and only fix $\beta$}
    \label{tab:psa-4}
\end{table}

\subsection{Ablation Study}
To validate the necessity of label distribution recovery modules into our method, we conduct ablation experiments on vector dataset Scene and image dataset PASCAL. Previous performance analysis shows that MulSupCon$_{\text{CLD}}$ outperforms MulSupCon$_{\text{RLD}}$ in overall performance. Therefore, we conduct ablation experiments on MulSupCon$_{\text{CLD}}$ and break it down into the following components: 
\begin{enumerate}[label=\arabic*)] 
    \item MulSupCon$_{\text{CLD}}^{w/o~\alpha}$: In this variant, we remove the constraints on the label distribution.
    \item MulSupCon$_{\text{CLD}}^{w/o~\beta}$: In this variant, we remove the term that controls the complexity of the label distribution model.
    \item MulSupCon$_{\text{CLD}}^{w/o~{all}}$: In this variant, we both remove the constraints on the label distribution and the term that controls the complexity of the label distribution model.
\end{enumerate}
We evaluate these methods on six metrics, and our results are illustrated in the Figure~\ref{fig:ablation-study}.

From the experimental results, we observe that removing any single component affects the model's performance. When we remove all components simultaneously, the model's performance drops significantly. We attribute this to two reasons: on one hand, the model tends to assign more description degrees to irrelevant labels; on the other hand, it becomes unable to effectively control the complexity of the label distribution model.

\subsection{Parameter Sensitivity Analysis}
In this section, we analyze the impact of the hyperparameters $\alpha$ and 
$\beta$ on the model's performance under the MulSupCon$_{\text{CLD}}$ method. 

We choose MulSupCon$_{\text{CLD}}$ because experimental performance analysis shows that it generally delivers better performance. We select both hyperparameters $\alpha$ and $\beta$ from the range \{0.001, 0.01, 0.1, 1\}. We fix hyperparameters $\alpha$ and $\beta$ separately and investigate their impact on the performance of MulSupCon$_{\text{CLD}}$ on both the vector dataset Scene and the image dataset PASCAL. We can find the results from Table~\ref{tab:psa-1} to Table~\ref{tab:psa-4}. From the results, we observed that for different datasets, the model required selecting different hyperparameters to achieve optimal performance. Additionally, on the Scene dataset, when we fixed hyperparameter $\alpha$ and increased hyperparameter $\beta$, the model eventually achieve its best performance. In contrast, on the PASCAL dataset, fixing hyperparameter $\beta$ and increasing hyperparameter $\alpha$ lead to the best performance for our model. We attribute this difference to the fact that image datasets are more complex than vector datasets. Increasing hyperparameter $\alpha$ helpe constrain the model to obtain better label distributions, thereby improving model performance.

\section{Conclusion}
We observe that current multi-label contrastive learning struggles to learn an effective model because using logical labels to balance classes losses overlooks the varying importance among labels. To address this problem, we propose MulSupCon$_\textbf{LD}$, a framework that improves multi-label contrastive learning by incorporating label distributions. In our method MulSupCon$_\textbf{LD}$, we recover label distributions from logical labels and utilize them to balance classes losses. Through the combined effects of contrastive learning and label distributions, our method demonstrates competitive performance across six metrics on nine datasets compared to comparative methods. However, multi-label learning commonly suffer from the long-tail distribution problem. Unlike the previous method MulSupCon, our method employs label distribution to directly capture dependencies between labels, enabling a better understanding and resolution of the long-tail distribution problem. Therefore, our future work will delve deeper into this area of research.


\acks{This research was supported by the National Natural Science Foundation of China (62306104), Hong Kong Scholars Program (XJ2024010), Jiangsu Science Foundation (BK20230949), China Postdoctoral Science Foundation (2023TQ0104), Jiangsu Excellent Postdoctoral Program (2023ZB140).}

\bibliography{sample}

\end{document}